\documentclass[conference]{IEEEtran}
\IEEEoverridecommandlockouts
\usepackage{cite}
\usepackage{amsmath,amssymb,amsfonts}
\usepackage{algorithmic}
\usepackage{graphicx}
\usepackage{textcomp}
\usepackage{multicol}
\usepackage{caption}
\usepackage{xspace}
\usepackage{multirow}
\usepackage{booktabs}

\usepackage{bm}

\usepackage{xcolor}
\def\BibTeX{{\rm B\kern-.05em{\sc i\kern-.025em b}\kern-.08em
    T\kern-.1667em\lower.7ex\hbox{E}\kern-.125emX}}
\begin{document}

\newtheorem{myDef}{Definition}
\newtheorem{exmp}{Example}
\newtheorem{remark}{Remark}

\newcommand{\paratitle}[1]{\vspace{1.5ex}\noindent\textbf{#1}}
\newcommand{\ie}{\emph{i.e.,}\xspace}
\newcommand{\aka}{\emph{a.k.a.,}\xspace}
\newcommand{\eg}{\emph{e.g.,}\xspace}
\newcommand{\etal}{\emph{et al.}\xspace}
\newcommand{\wrt}{\emph{w.r.t.}\xspace}

\newcommand{\sys}{EGL}
\newcommand{\procedure}{TRMP}
\newcommand{\rankmodel}{ALPC}




\title{Who Would be Interested in Services?
An Entity Graph Learning System for User Targeting\\

}

\author{

\IEEEauthorblockN{
Dan Yang \IEEEauthorrefmark{1}, Binbin Hu \IEEEauthorrefmark{1}, Xiaoyan Yang \IEEEauthorrefmark{1}, Yue Shen,
Zhiqiang Zhang,\\
Jinjie Gu \IEEEauthorrefmark{2},
Guannan Zhang \thanks{\IEEEauthorrefmark{1} Equal contributions } \thanks{\IEEEauthorrefmark{2} Corresponding author}
}
\IEEEauthorblockA{
\textit{Ant Group, China}\\
\{luoyin.yd, joyce.yxy\}@antgroup.com,\{bin.hbb, zhanying, lingyao.zzq, jinjie.gujj, zgn138592\}@antfin.com
}






}

\maketitle

\begin{abstract}
With the growing popularity of various mobile devices,  \emph{user targeting} has received a growing amount of attention, which aims at effectively and efficiently locating target users that are interested in specific services. Most pioneering works for \emph{user targeting} tasks commonly perform similarity-based expansion with a few active users as seeds, suffering from the following major issues: the unavailability of seed users for new-coming services and the unfriendliness of black-box procedures towards marketers.
In this paper, we design an \underline{E}ntity \underline{G}raph \underline{L}earning (EGL) system to provide explainable user targeting ability meanwhile applicable to addressing the cold-start issue. 
\textbf{\sys} System follows the hybrid online-offline architecture to satisfy the requirements of scalability and timeliness. 
Specifically, in the offline stage, the system focuses on the heavyweight entity graph construction and user entity preference learning, in which we propose a \underline{T}hree-stage \underline{R}elation \underline{M}ining \underline{P}rocedure ({\procedure}), breaking loose from the expensive seed users.
At the online stage, the system offers the ability of user targeting in real-time based on the entity graph from the offline stage. Since the user targeting process is based on graph reasoning, the whole process is transparent and operation-friendly to marketers. Finally, extensive offline experiments and online A/B testing demonstrate the superior performance of the proposed
{\sys} System. 
\end{abstract}


\section{Introduction}




The innovative mobile economy has served as a competitive market to provide internet companies (\eg Google, Tencent, and Alipay) with a variety of opportunities to promote their products and services.
Alipay has already become a platform for enabling inclusive, convenient digital life and digital financial services for consumers. 
Aiming at effectively and efficiently locating target users that are interested in certain services, \emph{user targeting}~\cite{mangalampalli2011feature,shen2015effective,ma2016score,dewet2019finding,zhuang2020hubble,zhu2021learning} has received a growing amount of attention, since its potential ability to derive high-quality users is well-aligned with marketers' needs for both facilitating the conversion population and reducing the operation costs. 

\begin{figure}[!t]
  \centering
  \includegraphics[width=0.9\columnwidth]{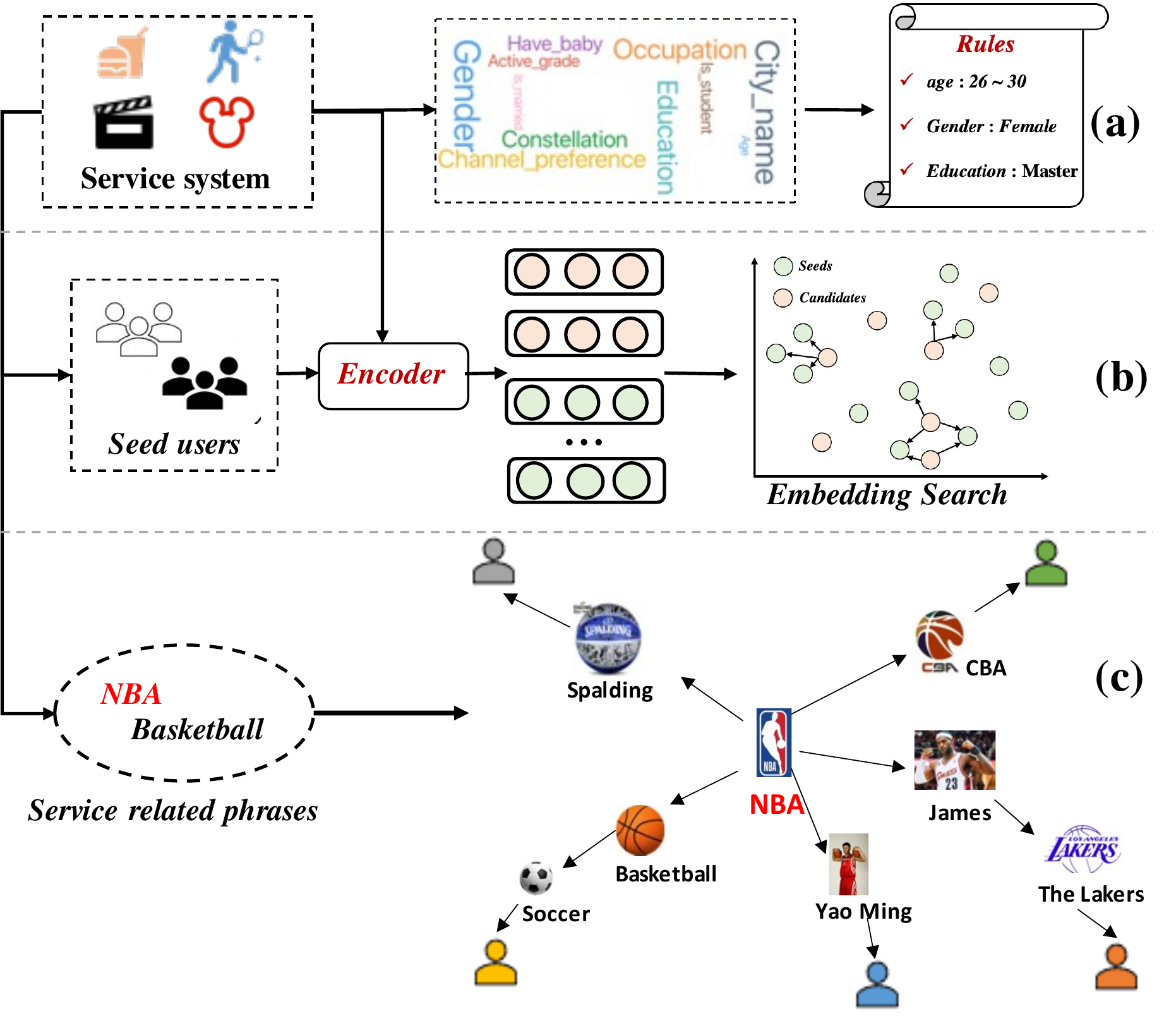}
  \caption{Comparison of three modes of user targeting }\label{fig:three_targeting_mode_comparison}
\end{figure}

Roughly speaking, current approaches devoted to \emph{user targeting} mainly fall into two lines.
The first type denotes the rule-based methods (Fig.~\ref{fig:three_targeting_mode_comparison} (a)) following the service-centered design, which targets users with prefabricated domain knowledge~\cite{mangalampalli2011feature,shen2015effective}, \ie tag mining and rule expression. 
As a comparison, the look-alike based methods (Fig.~\ref{fig:three_targeting_mode_comparison} (b)) seek to learn high-quality representations of seed users, and the target users can be effectively matched in the embedding space~\cite{zhu2021learning,ma2016score,dewet2019finding,zhuang2020hubble}. Owing to the powerful ability of representation learning for massive historical data summarization, the latter user-centered methods usually achieve better performance.
Unfortunately, they are still distant from optimal or even satisfactory in real scenarios, facing the following major issues:
i) New services appear every day, causing the unavailability of seed users for corresponding services. Besides, insufficient seed users may easily have coverage bias~\cite{zhuang2020hubble}. 
ii) The interpretability of user targeting is essential. Most look-alike based systems~\cite{zhu2021learning,ma2016score} utilize black-box algorithms to generate target user sets. 
Such an operation-unfriendly manner seems detrimental to the subsequent iteration of user targeting for marketers.

To fill this gap, we come up with a novel \underline{E}ntity
\underline{G}raph \underline{L}earning System (\textbf{\sys} System) for user targeting issues. 
As exhibited in Fig.~\ref{fig:three_targeting_mode_comparison} (c), given several phrases related to a specific service (\ie ``NBA" in Fig.~\ref{fig:three_targeting_mode_comparison} (c)), {\sys} System extends their connections iteratively along a well-established entity graph to discover their hierarchical relations (\eg ``NBA'' $\rightarrow$ ``James'' $\rightarrow$ ``The Lakers'' in Fig.~\ref{fig:three_targeting_mode_comparison} (c)). Based on the set of $k$-hop relevant entities, {\sys} System locates the target users with explicit preferences towards these candidate entities. 
{\sys} System performs cognitive reasoning upon entity graphs \wrt service-related phrases in an automatic manner, such that
i) service-based tag mining or seed users are not necessary for marketing the service, and; ii) entity graph based reasoning offers intuitive explanations for user targeting, as well as an interactive environment for marketers to flexibly control the depth of entity extension.

Constructing a high-quality entity graph is at the core of the {\sys} System, and given such a graph, it is critical to store and access relational knowledge of similar entities efficiently.
However, the entity relation mining process is non-trivial, given three intractable challenges:
i) The process of filtering undesired relations between entities is expected to be adaptive. ii) The negative sampling, the core of the model learning, is desired to be semantically augmented. iii) The 
entity relation mining procedure requires the stability of prediction due to the fluctuation of the data source.
To address these challenges, we propose a \underline{T}hree-stage \underline{R}elation \underline{M}ining \underline{P}rocedure ({\procedure}). In particular, we aim at gathering as many similar relations between entities as possible in the \emph{Candidate Generation Stage}, equipped with Skip-gram~\cite{mikolov2013efficient} based co-occurrence behaviors modeling and BERT~\cite{devlin2018bert} based semantic mining module. In terms of the subsequent \emph{Ranking Stage}, as the key component of {\procedure}, we endow the powerful ability of reliable relation filtering with \underline{A}daptive threshold \underline{L}ink \underline{P}rediction with \underline{C}ontrastive learning model ({\rankmodel}) based on graph neural networks \cite{scarselli2008graph,2017Inductive,2018Heterogeneous,2018GeniePath} and the semantic augmented contrastive strategy \cite{oord2018representation,jaiswal2020survey}. To maintain the stability of the entity mining process in the daily services, we present an \emph{Ensemble Stage} to integrate multiple entity representations derived from several well-trained ranking models with a multi-head attention encoder, such that the whole {\procedure} is more robust than one single ranking model.

To our knowledge, {\sys} System is the first user targeting system that automatically matches target users interested in services with efficient cognitive reasoning over well-established entity graphs, breaking loose from the expensive seed users and the black-box manner. We demonstrate the superiority of the proposed {\procedure} through extensive experiments on real-world datasets. Moreover, an in-depth analysis of online experiments also shows that our {\sys} System is effective, explainable, and operation-friendly. 
\section{System Overview}
\begin{figure}[!t]
  \centering
  \includegraphics[width=0.9\columnwidth]{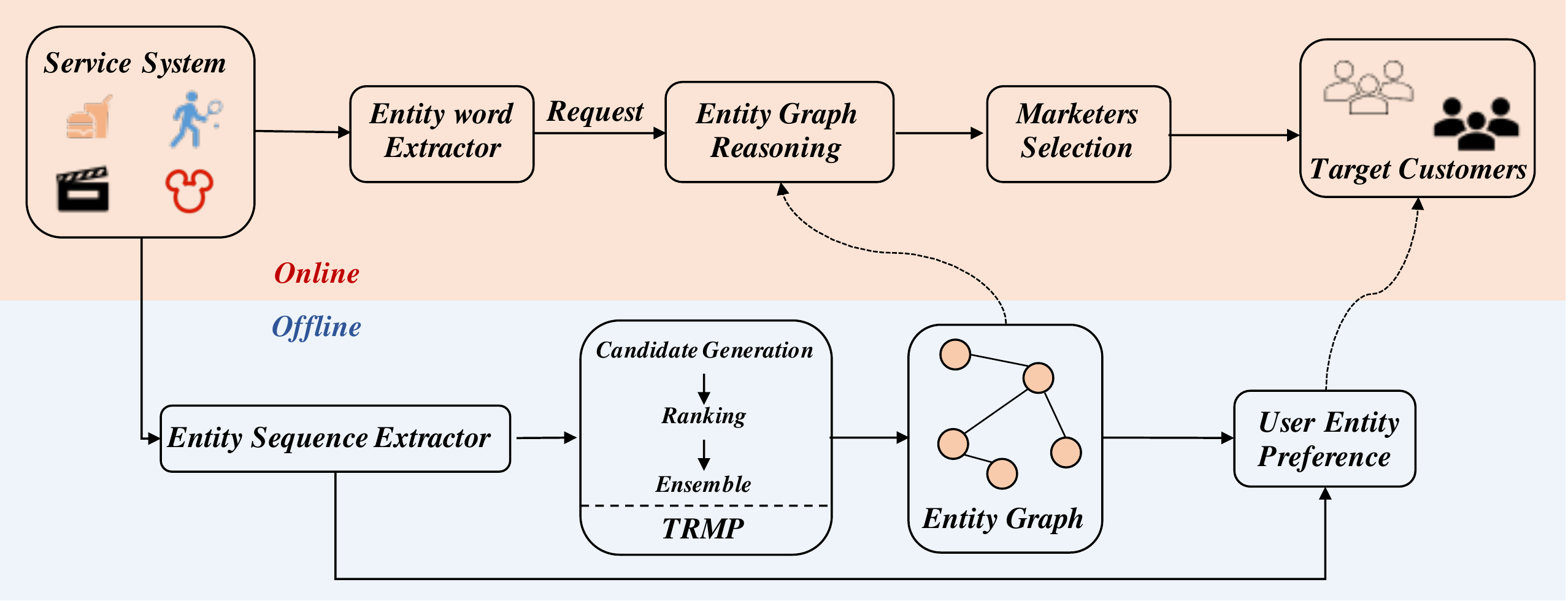}
  \caption{{\sys} System diagram, consisting of an offline pipeline and an online serving procedure.}\label{fig:CAT_system}
  \vspace{-8pt}
\end{figure}

In this section, we present the overview of the {\sys} System, following a hybrid online-offline architecture, shown in Fig.~\ref{fig:CAT_system}.


\subsection{Offline Stage}
The bottom part of Fig.~\ref{fig:CAT_system} shows the offline pipeline of the {\sys} System: \emph{Entity sequence extractor} $\rightarrow$ \emph{TRMP: Relation mining procedure} $\rightarrow$ \emph{Entity graph storage system} $\rightarrow$ \emph{User entity preference} (User preference generator towards entity). Specifically, the entity sequence extractor is responsible for collecting and preprocessing the data source (\eg user search and visit logs), which will be fed into the following relation mining procedure. Reliable relations between entities are fully mined through our proposed {\procedure} approach and further stored in the Alipay database called Geabase \cite{fu2017geabase}. Meanwhile, the entity embedding extracted in {\procedure} will be stored for the following module. To help {\sys} System locate target users effectively and efficiently in the online stage, the user entity preference module is employed to pre-compute user preference towards entities. Note that the offline stage is the cornerstone of supporting the online stage, the corresponding algorithm designs of each module will be detailed in Section~\ref{sec:model}.
\subsection{Online Stage}
The online stage of {\sys} System aims at discovering target users rapidly when a specific service needs to be promoted. In particular, a marketer is expected to request our {\sys} System with several phrases (represented as entities) related to the service. Centered on these given entities, the entity graph reasoning module extends their connections iteratively along the entity graph (well established in the offline stage) to discover their hierarchical potential relations. The depth of the extension could be flexibly controlled by marketers to achieve the trade-off between the relevancy and the diversity of the set of k-hop entities. Next, marketers select the entity they require and use the user entity preference module of the offline stage to retrieve all users associated with the chosen entities. Finally, given a central entity, {\sys} System only keeps top $K$ users with the highest average similarities, to whom the contents of the service will be promoted. 


\textbf{Remark}
{\sys} System runs 200 $\sim$ 300 user targeting experiments everyday. To keep the local structure and user entity preference up-to-date, the entity graph derived from the relation mining procedure is updated weekly and the user preference generator towards entity is in daily execution. Meanwhile, the relations chosen by marketers in the operation process will be recorded as relations with high confidence to guide the learning of the relation mining procedure, \ie our proposed {\procedure} framework.


\section{Diving into the offline stage of {\sys} system \label{sec:model}}
In this section, we will zoom into each module in the offline stage of {\sys} System.

\subsection{Entity Sequence Extractor}

\subsubsection{Entity Dict}
In real-world applications (\eg Alipay), users' behaviors are widely distributed throughout multiple scenarios, 
and the contents of different services are also diverse. To effectively employ user targeting over entity graphs, it is of crucial importance to perform content alignment between scenarios and services for entity-level uniformity. Hence, we introduce the \emph{Entity Dict} as the basis for bridging diverse contents and unified entities, each row of which is a tuple consisting of the \emph{entity} and \emph{entity type}. In particular, the Entity Dict is carefully designed by our dedicated group of experts, which involves millions of entities with 26 types.
It is worthwhile to note that the Entity Dict is automatically updated weekly to keep the fitness of entities. 


\subsubsection{Extracting Entities From Behaviors}
\begin{figure}[!t]
  \centering
  \includegraphics[width=1.0\columnwidth]{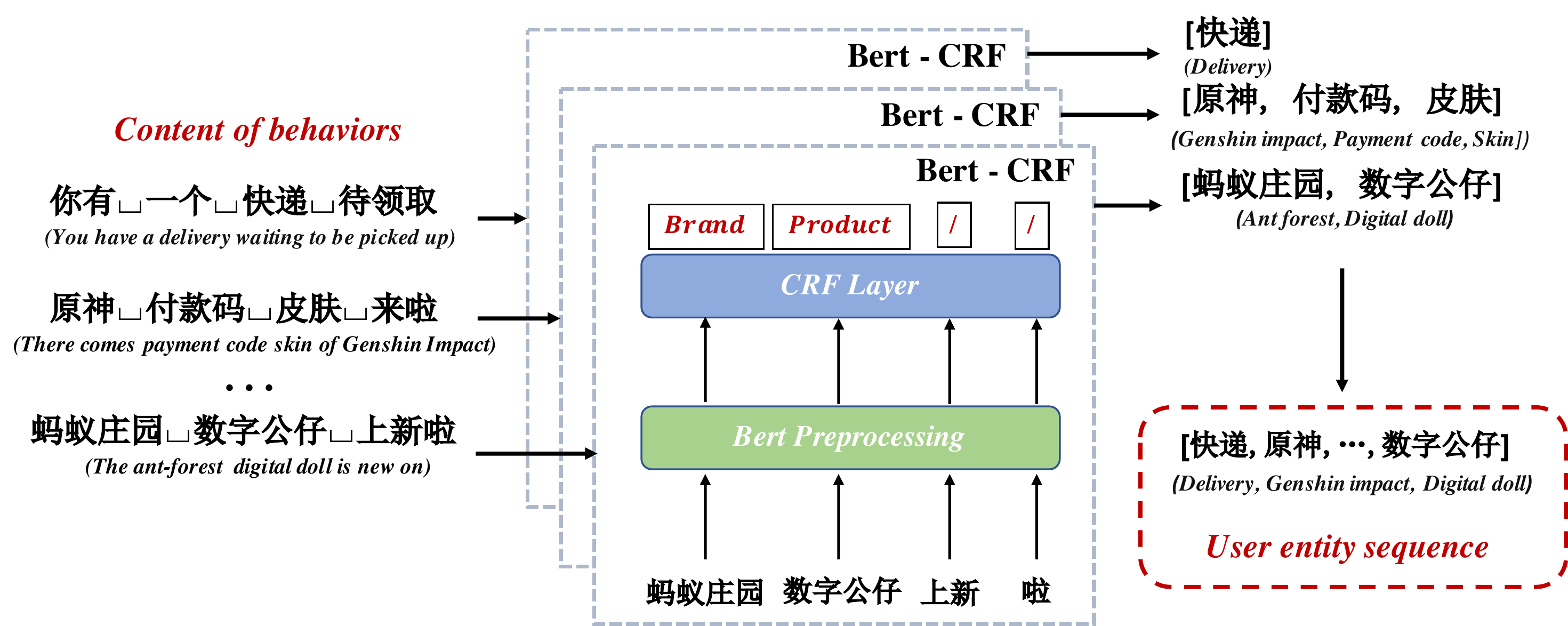}
  \caption{Extracting entities from behaviors}\label{fig:NER_Pipeline}
\end{figure}

Based on the \emph{Entity Dict}, we then shift attention to the entity extraction from a variety of user behaviors in Alipay, \eg search and visit logs. Naturally, such a process could be formulated as a NER task~\cite{zhang2018chinese,liu2019encoding,li2020survey}, which are widely studied in the NLP field. Hence, we adopt the state-of-the-art Bert\-CRF model to perform entity extraction, which combines the transfer capabilities of BERT~\cite{devlin2018bert} with the structured predictions of CRF~\cite{lafferty2001conditional}. For each user behavior, we feed the corresponding content into the Bert\-CRF~\cite{souza2019portuguese} model~\footnote{The Bert\-CRF model is well pre-trained based on manually labeled data.}, whose output is an entity list tagged on the user behavior. Moreover, we collect user behaviors in the past 30 days, which conducts the final entity sequence via chronological concatenation. The overall procedure is detailed in Fig.~\ref{fig:NER_Pipeline}.

\subsection{{\procedure} Design} 
\begin{figure*}
  \centering
\includegraphics[width=1.8\columnwidth]{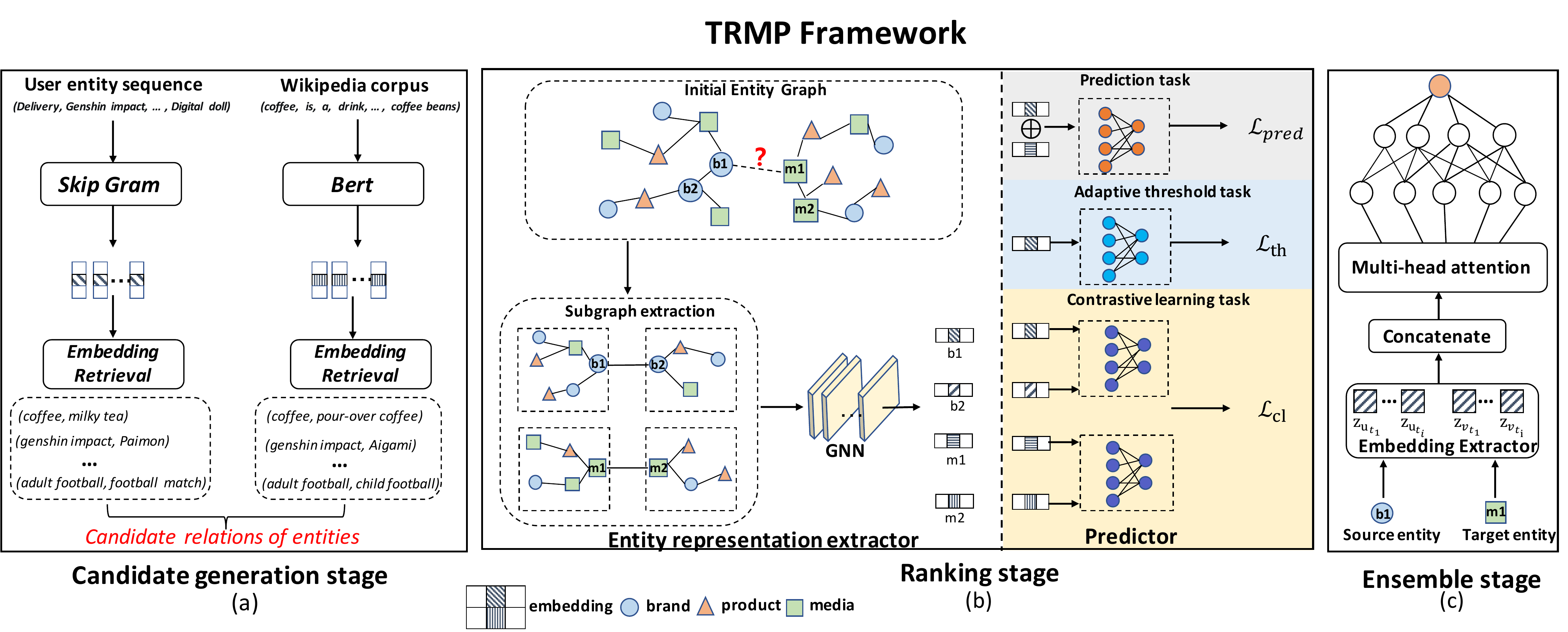}
  \caption{Overview of \textbf{{\procedure}} framework consisting of three stages. 
  }\label{fig:TSEEMP}
\end{figure*}

Intuitively, the success of the {\sys} System greatly hinges on the building of entity graph with high quality, and thus, we propose the Three-stage Relation Mining Procedure, called {\procedure}, which consists of the \textbf{candidate generation}, \textbf{ranking} and the \textbf{ensemble stage}, as shown in Fig.~\ref{fig:TSEEMP}. In the following parts, we will take a closer look at each well-designed stage.

\subsubsection{Stage I: Candidate generation}
As shown in Fig.~\ref{fig:TSEEMP} (a), the candidate generation task aims to generate the initial entity graph from co-occurrence and semantic aspects.
In particular, we adopt the Skip-gram model~\cite{mikolov2013efficient} to mine the co-occurrence relevance between entities among the abundant entity sequences derived from the entity sequence extractor.
In terms of the semantic-level relevance, we utilized the Bert~\cite{devlin2018bert}, which is pre-trained on a large amount of public corpus, \eg Wikipedia ~\footnote{https://dumps.wikimedia.org/zhwiki/}.
Moreover, we denote the graph in this stage as $\mathcal{G}^{C}$, and respectively denote the co-occurrence-level and semantic-level embedding matrices as $\mathbf{E}^{Se}$ and $\mathbf{E}^{Co}$, which will be used in the ranking stage.

Through manual evaluation of the relations generated from the candidate generation stage, we surprisingly find the accuracy of all the relations far lower than 90\%. A fine-grained ranking stage is necessary to improve the accuracy.



\subsubsection{Stage II: Ranking Stage}
\begin{figure}[!t]
  \centering
  \includegraphics[width=1.0\columnwidth]{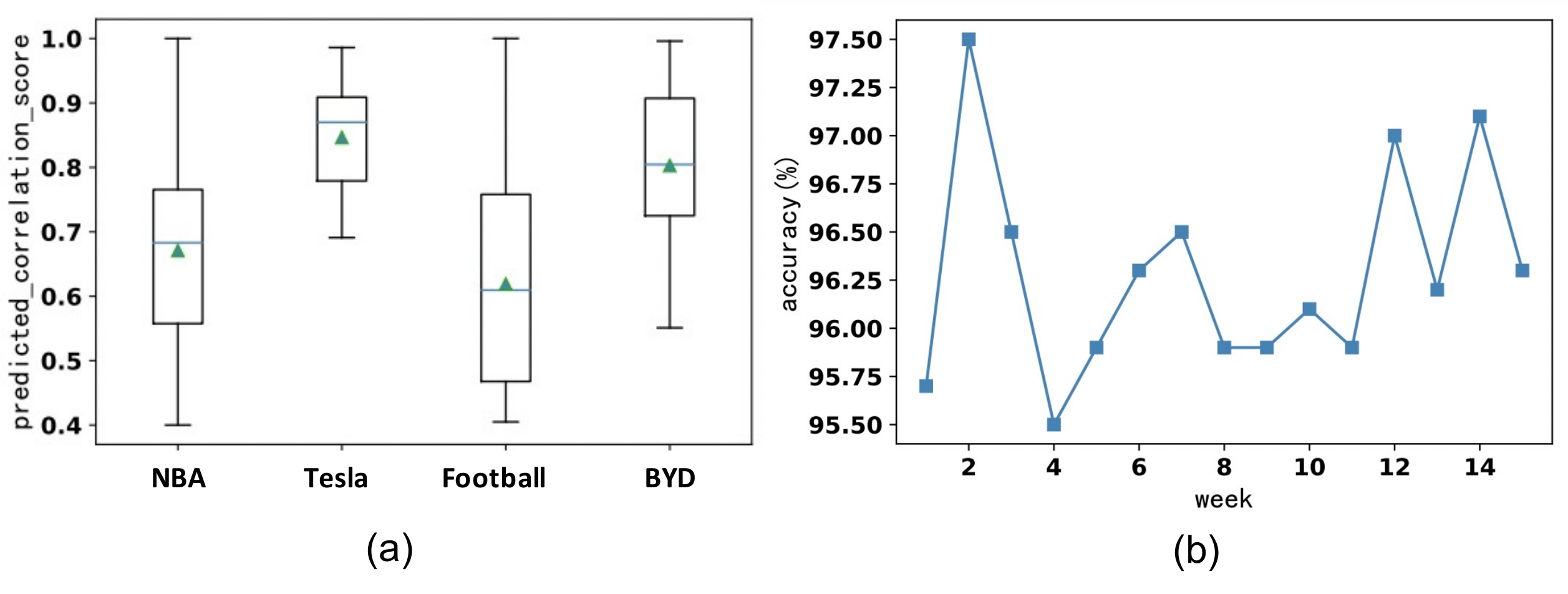}
  \caption{(a) Skewed distribution of predictions \wrt different source entities. (b) The weekly accuracy trend of ALPC.}\label{fig:boxplot_badcase}
    \vspace{-10pt}
\end{figure}

The performance of the ranking stage greatly hinges on the correlated entity pairs retrieved in the candidate generation stage. Generally, it can be formulated as a link prediction task~\cite{liben2003link,adamic2003friends,katz1953new,tang2015line,perozzi2014deepwalk,grover2016node2vec,shi2018heterogeneous,zhang2018link,teru2020inductive}, which could improve the accuracy of the existing relations derived from the candidate generation stage, as well as explore unknown relations for the richness of the target entity graph.
As a powerful tool for exploiting structural information, graph neural networks~\cite{scarselli2008graph,2017Inductive,2016Semi,2017Graph,2018GeniePath,bo2022regularizing} have been widely applied in link prediction tasks\cite{zhang2018link,teru2020inductive}, and attain great success. Due to its excellent performance, we adopt the GeniePath~\cite{2018GeniePath} as the backbone for entity representation. Formally, given a source and target entity pair $(u, v)$, the semantic-level and co-occurrence-level embedding \ie \ \{$\bm{e}_u^{Se}$, $\bm{e}_v^{Se}$\} and \{$\bm{e}_u^{Co}$, $\bm{e}_v^{Co}$\} (element of $E^{Se}$ and $E^{Co}$) will be fed into Geniepath as entity features, 
 the whole encoding process is as follows:
\begin{equation}
    \bm{z}_u = f_{GeniePath} ([\bm{e}_u^{Se},\bm{e}_u^{Co}]), \\
    \bm{z}_v = f_{GeniePath} ([\bm{e}_v^{Se},\bm{e}_v^{Co}]).
\end{equation}
Based on the representation, a graph neural network based link prediction could be well-trained through the widely-adopted CrossEntropy-based objective~\cite{rubinstein2004cross,zhang2018link,hasan2011survey}.
\begin{equation}
    \begin{split}
         s_{u,v} &= g([\bm{z}_u||\bm{z}_v]), \ \ \ \ \hat{y}_{u,v} = \sigma(s_{u,v}), \\
         \mathcal{L}_{pred} &= - \sum y_{u, v}log(\hat{y}_{u, v}) + (1-y_{u, v})log (1-\hat{y}_{u, v}),
    \end{split}
\end{equation}
where $g(\cdot)$ is a scoring function (e.g., inner product, bilinear function or a neural network), $y_{u,v}$ is the ground truth and $\hat{y}_{u,v}$ is the predicted correlation score between the source entity and target entity. Nevertheless, employing such an optimization procedure in our scenarios still faces the following challenges: 

\textbf{Challenge 1}: Different source entities have different correlated target entities, and the distribution of the predicted correlation scores $\hat{y}_{u,v}$ of each source entity is different, shown in Fig.~\ref{fig:boxplot_badcase} (a), where NBA's score distribution is similar to football's while Tesla's score is similar to BYD's.
Hence, when we make threshold truncation, the threshold should be different for different source entities. 


\textbf{Challenge 2}: Previous research in metric learning has established that the hard negative sample is of particular concern in representation learning, while traditional link prediction methods commonly adopt the native random sampling strategy, such that derived ``easy'' samples are prone to restrict the performance \cite{oord2018representation,schroff2015facenet}.


So we propose a novel link prediction model \textbf{{\rankmodel}} to tackle both challenges. As seen in the ranking stage of Fig.~\ref{fig:TSEEMP} (b), {\rankmodel} mainly adds an adaptive threshold task and a contrastive learning task to the former optimization procedure, aiming at handling challenge 1 and challenge 2 respectively.

\textbf{Adaptive threshold learning task} The task is to learn the personalized threshold of each source entity. So we add Multi Layer Perceptron (MLP) to process the source entity and predict a threshold score $\epsilon$.
Aiming at enlarging the margin between prediction score $s$ and threshold $\epsilon$, the loss of adaptive threshold learning task is defined as follows, 

\begin{equation}
    \begin{split}
    \epsilon_u & = \text{MLP}(\bm{z}_u), \ \ \ \ \hat{y}'_{u,v} = \sigma(s_{u,v} - \epsilon_{u}), \\
    \mathcal{L}_{th} &= - \sum y_{u, v}log(\hat{y}'_{u, v}) + (1-y_{u, v})log (1-\hat{y}'_{u, v}).
    \end{split}
\end{equation}


\textbf{Contrasive learning task}
Inspired by contrastive learning~\cite{oord2018representation,le2020contrastive,jaiswal2020survey}, which works by pulling positive samples closer and pushing negative samples further, we enhance the representation of entities with the auxiliary contrastive supervision. In real-world applications, entities are commonly associated with abundant textual information, which is a beneficial signal for facilitating representation learning. In particular, for a (source or target) entity $e$, we construct the anchor pairs $<e, e+>$ through semantic-level similarities higher than a threshold in all the correlated entity lists. 
Subsequently, our contrastive learning objective is to minimize the following function based on InfoNCE~\cite{oord2018representation,2010Noise}:
\begin{equation}
    \mathcal{L}_{cl} = \sum{\text{log} \frac{\text{exp}(\bm{z}_e \cdot \bm{z}_{e^+} / \tau)}{\sum_{e^{-}}{\text{exp}(\bm{z}_e \cdot \bm{z}_{e^-} / \tau)}}},
\end{equation}
where $\tau$ is the temperature hyper-parameter 
and $e^-$ is drawn from the widely-used in-batch negative sampling strategy.

The model's total loss is the weighted sum of prediction loss $\mathcal{L}_{pred}$, threshold loss $\mathcal{L}_{th}$, and contrastive loss $\mathcal{L}_{cl}$ with hyper-parameters $\alpha$ and $\beta$. Experimentally, our model yields the best performance when $\alpha = \beta = 1$. 

\begin{equation}
\mathcal{L}=\mathcal{L}_{pred} + \alpha * \mathcal{L}_{th} + \beta*\mathcal{L}_{cl}.
\end{equation}

\subsubsection{Stage III: Ensemble Stage}

Due to the change in data distribution of upstream data sources (search logs, visit logs, etc.), the ranking model {\rankmodel}'s performance is not stable enough, thus resulting in a big fluctuation of accuracy.
Fig.~\ref{fig:boxplot_badcase} (b) shows the weekly accuracy trend of {\rankmodel}. The accuracy’s upper bound is up to 97.5\% and the lower bound is 95.5\%, where the accuracy's variance is up to 0.31.


Therefore, we add the ensemble stage (Fig.~\ref{fig:TSEEMP} (c)) to improve the stability of the accuracy. It consists of an embedding extractor, a multi-head attention encoder, and MLP modules. Since the {\rankmodel} model is updated weekly, we can extract the entity embedding $\bm{z}_{e_{t_{i}}}$ from weekly ALPC model respectively.

\begin{equation}
\begin{aligned}
\bm{h}_{u}&=Concatenate(\bm{z}_{u_{t_{1}}}, ..., \bm{z}_{u_{t_{i}}}), \ i=1,2,...
    \\
\bm{h}_{v}&=Concatenate(\bm{z}_{v_{t_{1}}}, ..., 
\bm{z}_{v_{t_{i}}}), \ i=1,2,...
 \\
 \ \bm{H}_{u, v} &=Concatenate(\bm{h}_{u}, \bm{h}_{v}).
 \end{aligned}
\end{equation}
And new predicted values will be obtained through multi-head attention encoder and MLP module. We also use cross entropy to calculate the loss of this stage. Specifically, we store the concatenated entity embedding $\bm{h}_e$ for the following module.




\subsection{Entity Graph Storage and User Entity Preference}
\textbf{Entity Graph Storage.} The relations mined from {\procedure} framework such as $<$NBA, CBA$>$, would form the final entity graph and we store it in the Geabase for online serving~\cite{fu2017geabase}. 
 
\textbf{User Entity Preference.} The inputs of this module are user entity sequence from the entity sequence extractor module and entity embedding $\bm{h}_e$ extracted from the ensemble task. The user embedding is the element-wise sum of $\bm{h}_e$ in the corresponding user entity sequence. The dot product between user embedding and entity embedding is the user entity preference score. The equations are as follows. $r_{u_k}$ denotes the embedding of user $k$, $l$ denotes the length of user entity sequence, $\bm{s}_{<u_k,e_m>}$ denotes the user's preference score towards entity ${m}$.

\begin{equation}
\begin{aligned}
\bm{r}_{u_{k}}= \sum_{j=1}^l \bm{h}_{e_j}/l, 
\bm{s}_{<u_k,e_m>}= \bm{r}_{u_k} * \bm{h}_{e_m}.
\end{aligned}
\end{equation}

\section{Experiments}


\subsection{Experiment Settings}


\subsubsection{Evaluation metrics} 
To evaluate the key components of {\sys} System, \ie {\procedure} and {\rankmodel}, we adopt several metrics, including ACC (Accuracy), CorS (Correlation Score), AEEC (Average Expansion Entity Count), AUC (Area under the ROC Curve), where ACC and CorS are calculated 
through manual evaluation while AEEC denotes the average number of correlated entities for each source entity.
\textbf{Details of manual evaluation}: we randomly sample entity pairs and ask 8 annotators to decide whether the entity pairs are correlated. The annotators have three choices: highly correlated, medium correlated, and uncorrelated, which respectively represent correlation score=1, 0.5, 0. And ``correlation score = 0 " denotes inaccurate relation while ``correlation score $>$ 0 " denotes accurate relation.
\begin{equation}
\begin{aligned}
CorS=\frac{\sum_{i=1}^N \sum_{j=1}^N C_{i,j}}{\sum_{i=1}^N \sum_{j=1}^N T_{i,j}},
AEEC= \frac {\sum_{i=1}^N \sum_{j=1}^N T_{i,j} }{N},
\end{aligned}
\end{equation}
where $N$ is the amount of the Entity Dict, $C_{i,j}$ represents the correlation score and $T_{i,j}$ denotes whether there is a relation between entities.





\subsubsection{Datasets}

Since this paper mainly focuses on the industrial problem
of user targeting of service in the digital marketing scenario, we employ the real-world industrial datasets~\footnote{The data set does not contain any Personal Identifiable Information (PII). The data set is desensitized and encrypted. Adequate data protection was carried out during the experiment to prevent the risk of data copy leakage, and the data set was destroyed after the experiment.} of Alipay. And for different stages of {\procedure}, the dataset differs.

\paratitle{Dataset of candidate generation stage} The dataset of the co-occurrence part is the user entity sequence which comes from the entity sequence extractor, consisting of about ten million samples after random sampling 
The dataset of the semantic part comes from the public Wikipedia corpus.

\paratitle{Dataset of ranking and ensemble stage} Since the {\procedure} framework includes manual evaluation after the candidate generation stage, we retain the relations of the candidate generation stage only if the accuracy achieves a certain threshold and form the initial entity graph. The initial entity graph has millions of entities and billions of edges, making up 78\% of the total entity dictionary.
We randomly remove 10\% of existing relations from the initial graph as positive testing data. Following a standard manner of learning-based link prediction, we randomly sample the same number of nonexistent relations (unconnected node pairs) as negative testing data. We use the
remaining 90\% of existing links as well as the same number of additionally sampled nonexistent links to construct the training data. The other negative samples come from negative sampling methods. In short, the datasets of both ranking and ensemble stage consist of 6 million positive samples and 18 million negative samples, called \textbf{Dataset-M}. 


\subsection{Effectiveness of {\procedure}}
The performance of each stage of {\procedure} is reported in Table~\ref{tab:metrics}, in which we prepare two variants of {\procedure}, \ie \ {\procedure} w.o E without ensemble stage and {\procedure} w.o E\&R without both ensemble and ranking stage. And the stage TRMP w.o. E\&R$_s$ denotes forming entity pairs through popularity sampling methods from Entity Dict. We can observe that in terms of ACC and CorS metrics, {\procedure} $>$ {\procedure} w.o E $>$ {\procedure} w.o E\&R $>$ TRMP w.o. E\&R$_s$. The candidate generation stage improves the ACC from 68.6\% to 80.6\%, and the ranking stage improves the ACC from 80.6\% to 97.7\%. We can obtain that the ranking stage plays the most important role in terms of ACC and CorS metrics. The AEEC metric of the candidate generation stage is the biggest, which shows the richness of expanding entities. In terms of the variance of ACC, the ensemble stage shows great potential (0.31 $\rightarrow$ 0.08).

In summary, the whole {\procedure} framework improves the ACC to 97\%+ through the ranking stage and maintains the ACC \& CorS in a steady value through the ensemble stage, which achieves an ideal level for entity graph construction.

\begin{table}[tp]
  \setlength\tabcolsep{2pt}
\caption{Metrics of each stage}
  \centering
  
    \begin{tabular}{ccccc}
      \hline
      Stage &{ ACC}&{CorS} &{AEEC} & {Variance of ACC}\\
      \hline
      { {TRMP w.o. E\&R}$_s$}  & 68.60\%  & 0.673 & 78.0 & 0.30 \\
      
      { {\procedure} w.o E\&R } & 80.60\% & 0.780 & 78.0 & 0.32 \\ 
      
      { {\procedure} w.o E} & 97.70\% & 0.950  & 61.2 & 0.31\\ 
    {{\procedure} }& 97.76\% & 0.951  & 59.5 & 0.08\\
    \hline
    
\end{tabular}
\label{tab:metrics}
\vspace{-10pt}
\end{table}


\subsection{Effectiveness of {\rankmodel}}

\textbf{Baseline methods} In this experiment, we prepare the following baselines:
(1) Graph embedding based methods: DeepWalk \cite{perozzi2014deepwalk}, Node2Vec \cite{grover2016node2vec}. (2) GNN-based methods: VGAE \cite{kipf2016variational}, SEAL\cite{zhang2018link}, Geniepath \cite{2018GeniePath}, CompGCN \cite{2019Composition}, PaGNN \cite{yang2021inductive}. 
To verify the consistent performance of {\rankmodel}, We evaluate {\rankmodel} and other methods on three sampled sub-dataset A, B, and C with different sampling ratios (details can be seen in Table~\ref{tab:offline}) from Dataset-M, and report the results in Table~\ref{tab:offline}.
In addition, to verify the effectiveness of the proposed auxiliary tasks, we prepare variants ${\rankmodel}_{th-}$ (\ie \ {\rankmodel} without the adaptive threshold network), and ${\rankmodel}_{cl-}$ (\ie \ {\rankmodel} without the contrastive learning task).


From Table~\ref{tab:offline}, we have the following observations and analyses.
First, among all the methods, {\rankmodel} performs best in all the datasets for both metrics, especially the ACC (manual evaluation metrics), indicating the effectiveness of {\rankmodel} by bringing in the adaptive threshold network and contrastive learning task.
Second, we compare {\rankmodel}, ${\rankmodel}_{th-}$ and ${\rankmodel}_{cl-}$. The AUC gap between {\rankmodel} and ${\rankmodel}_{th-}$ is low since the latter version only lacks an adaptive threshold task, which mainly learns a threshold score. However, the {\rankmodel}'s ACC is much better than ${\rankmodel}_{th-}$, which verifies the effectiveness of the adaptive threshold task. The comparison between {\rankmodel} and ${\rankmodel}_{cl-}$ shows that adding a contrastive learning task can improve the ACC greatly. On the other hand, we find that the contrastive learning task is better than the adaptive threshold task in improving ACC.


\begin{table}[tp]
  \caption{Performance comparison on offline datasets.}
    \begin{tabular}{c|cc|cc|cc}
      \toprule 
        {} & \multicolumn{2}{c|}{Dataset A } & \multicolumn{2}{c|}{ Dataset B } & \multicolumn{2}{c}{Dataset C} \\
      {\# Entities}  & \multicolumn{2}{c|}{113,267} & \multicolumn{2}{c|}{42,529} & \multicolumn{2}{c}{92,651} \\
      {\# Edges}  & \multicolumn{2}{c|}{11,570,856 } & \multicolumn{2}{c|}{4,337,924} & \multicolumn{2}{c}{9,272,733 } \\
     \midrule
     \midrule
      Methods & AUC & ACC  &AUC&ACC&AUC&ACC\\
      \midrule
      {DeepWalk} & 0.846  &  0.909 & 0.837  & 0.911 & 0.852 & 0.921\\ 
      {Node2Vec} & 0.848  &  0.915 & 0.839 & 0.913 & 0.856 & 0.932\\ 
      \midrule 
      {SEAL} & 0.868  &  0.940 & 0.863 & 0.936 & 0.873 & 0.943\\ 
      {VGAE} & 0.847 &  0.928  &0.857 & 0.930 & 0.874 & 0.939\\ 
      {Geniepath} & 0.870 & 0.944 & 0.865 & 0.942 & 0.877 & 0.945 \\
      CompGCN & 0.869 & 0.942 & 0.865 & 0.943  & 0.876 & 0.944 \\ 
      PaGNN & 0.872 & 0.951 & 0.867 & 0.951 & 0.878 & 0.955 \\ 
      \midrule 
      { $\rankmodel$ } & \textbf{0.879}  &  \textbf{0.967}  &  \textbf{0.870} &  \textbf{0.961} &  \textbf{0.883} &  \textbf{0.973} \\ 
      \midrule
      
      { {\rankmodel}$_{th-}$ } & 0.875 & 0.960 & 0.868 & 0.956& 0.882 & 0.960  \\ 
      { {\rankmodel}$_{cl-}$ } & 0.871 & 0.950 & 0.862 & 0.944  & 0.879 & 0.953 \\ 
      
    \bottomrule
\end{tabular}

  \label{tab:offline}
\end{table}

\subsection{Online Performance of the {\sys} System}
The {\sys} System has already been deployed in the production environment of Alipay to serve the marketers to promote their services. Here, we conduct online A/B testing experiments to demonstrate the performance of the {\sys} System (shown in Table~\ref{tab:online}) in real traffic when there are no seed users of the service. And we report the results based on the following four metrics: \# \textbf{exposure} means the number of users who have been exposed by the service, \# \textbf{conversion} means the number of users who have clicked inside the service, \textbf{CVR} means the conversion rate of the service, \textbf{and running time} means the total running time of user targeting task. 

\textbf{Effectiveness} Note that the user targeting task is expected to find users who are most likely like the service. A higher CVR indicates a higher quality of selected users, 
and we report the gains of the {\sys} against the online baseline (\ie rule-based method) in Table~\ref{tab:online}. It shows that the proposed {\sys} System maintains a great improvement in conversion and CVR. 

\textbf{Efficiency} In terms of user targeting efficiency, the whole operation process only needs 2-4 minutes on average, which is 3 times faster than the former system in Alipay, \ie Hubble System \cite{zhuang2020hubble}.
In summary, the proposed {\sys} System performs effectively and efficiently in real-world online A/B testing, thus being more suitable for the industry.


\begin{table}[tp]
  \centering
  \setlength\tabcolsep{2pt}
  \caption{Online experiments performance}
  \vspace{0.1cm}
    \begin{tabular}{lcccc}
    \hline
      Services &{ \# exposure }&{ \# conversion} &{CVR} & {Running Time}\\
      \hline
      {Railway} & +0.30\%  & 23.20\% & 23.00\%  & 3.0 min \\ 
      {Dicos} & +0.50\% & 16.90\%   & 16.30\%  & 2.0 min\\ 
    {Cosmetics}& -0.20\% & 19.50\%   & 19.80\% & 2.5 min\\
    {Dessert}& +0.73\% & 33.60\%   & 32.90\%  & 3.2 min\\
    {Women Football}& +0.10\% & 9.40 \%  & 9.20 \%  & 2.2 min\\
    \hline
\end{tabular}
  \label{tab:online}
\vspace{-10pt}
\end{table}

\subsection{Application Case}

\begin{figure}[!t]
  \centering
  \includegraphics[width=0.9\columnwidth]{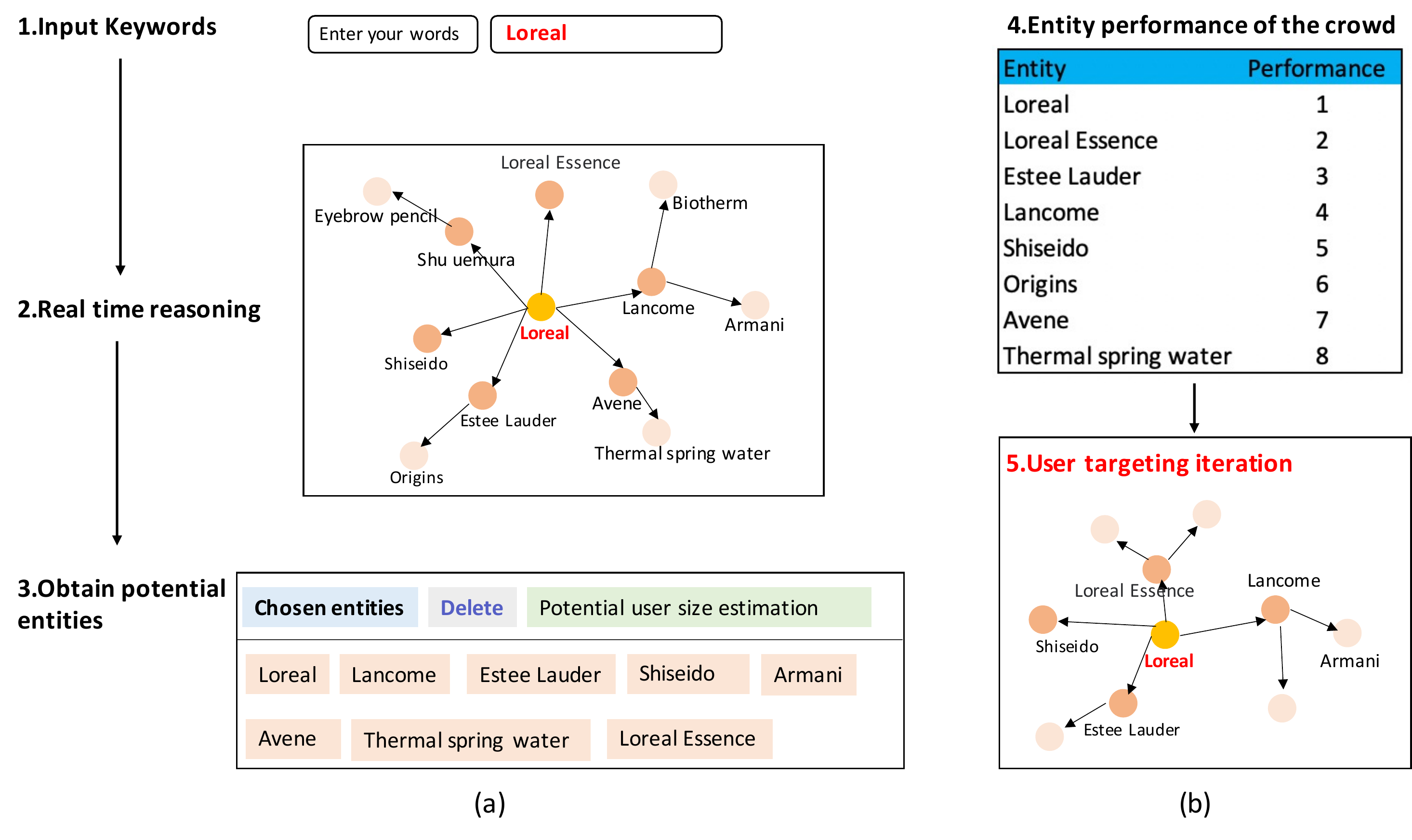}
  \caption{A real case of user targeting of {\sys} system in Alipay
}\label{fig:audience_targeting_case}
\end{figure}

In this part, we will show a practical application case of the {\sys} System including processes of both user targeting and the marketer's next iteration to the target users. Fig.~\ref{fig:audience_targeting_case} (a) and Fig.~\ref{fig:audience_targeting_case} (b) represent these two processes respectively. In the user targeting process, when a marketer brings in a new service namely the L'Oreal service on the Alipay app, 
the marketer needs to propagandize it. In this scenario, the marketer only needs to search the word L'Oreal (entity) in the input box (the first step in Fig.~\ref{fig:audience_targeting_case} (a)) and our system will show the entity and the entity's two-hops subgraph in default (the second step in Fig.~\ref{fig:audience_targeting_case} (a), the marketers can choose any hops they need). Then they choose the entities they need and the chosen entities will be at the bottom of the final box (the third step in Fig.~\ref{fig:audience_targeting_case}). When the marketer clicks the export button, our {\sys} will compute the target users of certain services and complete the exportation. The whole user targeting process only needs 2-4 minutes on average. Once the target users are exported and used for the service, the {\sys} computes the performance of all the chosen entities of corresponding target users (the fourth step in Fig.~\ref{fig:audience_targeting_case}). Note the performance, the marketers can promote their service's performance by iterating the user targeting process. Hence through our system, the marketers can not only obtain the service's target users on their own,
but also they can iterate the above process to discover the target users that meet their needs.

\section{Conclusion}
In this paper, we propose an innovative industrial system for audience targeting in mobile marketing scenarios, called {\sys} System, supported by our well-designed {\procedure} framework. In particular, the {\procedure} framework consists of candidate generation, ranking and ensemble stage, where we proposed a novel {\rankmodel} for effective relation mining in the ranking stage. Extensive experiments in offline and online environments demonstrate the effectiveness and efficiency of {\sys} System.

\paratitle{Future works} 
The dynamic nature of entity graphs renders ALPC vulnerable to the distribution shift in realistic scenarios, thus incorporating stable learning~\cite{shen2020stable,zhang2021deep} and causal inference~\cite{wu2022discovering} for out-of-distribution generalization is a promising direction. Moreover, we are also interested in investigating hyperbolic graph learning~\cite{liu2019hyperbolic,yang2022hyperbolic} for modeling hierarchical structures in our entity graphs.

\bibliographystyle{IEEEtran}
\bibliography{sample_base} 


\end{document}